\pdfoutput=1
\documentclass[11pt]{article}

\usepackage[preprint]{acl}

\usepackage{times}
\usepackage{latexsym}
\usepackage{booktabs}
\usepackage[T1]{fontenc}
\usepackage[utf8]{inputenc}

\usepackage{microtype}

\usepackage{inconsolata}
\usepackage{enumitem}
\usepackage{graphicx}
\usepackage{tabularx}
\usepackage{array}
\usepackage{ragged2e}
\usepackage{pifont}
\usepackage{subcaption}
\usepackage{amsmath}
\usepackage{amssymb}
\newcolumntype{Y}{>{\Centering\arraybackslash}X}
\usepackage[labelsep=space]{caption}

\usepackage{tikz}
\newcommand*\circled[1]{\tikz[baseline=(char.base)]{
            \node[shape=circle,draw,inner sep=0.4pt] (char) {#1};}}
            
\definecolor{codegreen}{rgb}{0.0, 0.411, 0.243}
\definecolor{codered}{rgb}{0.89, 0.26, 0.20}
\definecolor{custompurple}{RGB}{165,119,233}
\definecolor{customblue}{RGB}{0,102,204}
\usepackage{hyperref}
\definecolor{dartgreen}{HTML}{00693e}
\definecolor{refcolor}{HTML}{9F363A}

\hypersetup{
    colorlinks=true,
    linkcolor=refcolor,
    citecolor=refcolor,
    filecolor=magenta,      
    urlcolor=refcolor,
    }
\title{SoundMind: RL-Incentivized Logic Reasoning for Audio-Language Models}

\author{
 \textbf{Xingjian Diao}$^\heartsuit$,
 \textbf{Chunhui Zhang}$^\heartsuit$,
 \textbf{Keyi Kong} $^\diamondsuit$,
 \textbf{Weiyi Wu}$^\heartsuit$,
 \textbf{Chiyu Ma}$^\heartsuit$,
   \\
 \textbf{Zhongyu Ouyang}$^\heartsuit$\textbf{,}
  \textbf{Peijun Qing}$^\heartsuit$\textbf{,}
 \textbf{Soroush Vosoughi}$^\heartsuit$\textbf{,}
 \textbf{Jiang Gui}$^\heartsuit$
\\
 $^\heartsuit$Dartmouth College,  $^\diamondsuit$Shandong University
 \\
   \texttt{xingjian.diao.gr@dartmouth.edu}
 \\
\raisebox{-0.25em}{%
  \includegraphics[width=0.033\linewidth]{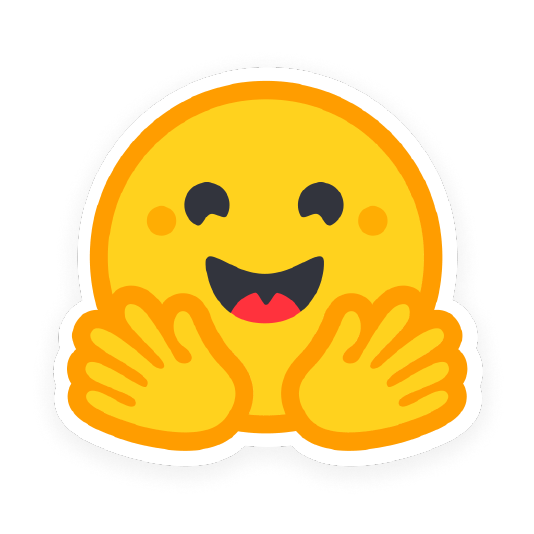}%
}%
{\hypersetup{urlcolor=custompurple}\href{https://huggingface.co/datasets/SoundMind-RL/SoundMindDataset}{SoundMind Dataset}}
}
\begin{document}
\maketitle

\begin{abstract}
While large language models have demonstrated impressive reasoning abilities, their extension to the audio modality, particularly within large audio-language models (LALMs), remains underexplored. Addressing this gap requires a systematic approach that involves a capable base model, high-quality reasoning-oriented audio data, and effective training algorithms. In this work, we present a comprehensive solution for audio logical reasoning (ALR) tasks: we introduce SoundMind, a dataset of 6,446 audio–text annotated samples specifically curated to support complex reasoning. Building on this resource, we propose SoundMind-RL, a rule-based reinforcement learning (RL) algorithm designed to equip audio-language models with robust audio–text reasoning capabilities. By fine-tuning Qwen2.5-Omni-7B on the proposed SoundMind dataset using SoundMind-RL, we achieve strong and consistent improvements over state-of-the-art baselines on the SoundMind benchmark. This work highlights the benefit of combining high-quality, reasoning-focused datasets with specialized RL techniques, and contributes to advancing auditory intelligence in language models.
The code and dataset are publicly available at  \url{https://github.com/xid32/SoundMind}.
\end{abstract}

\section{Introduction}

\begin{figure*}[t]
  \includegraphics[width=\textwidth]{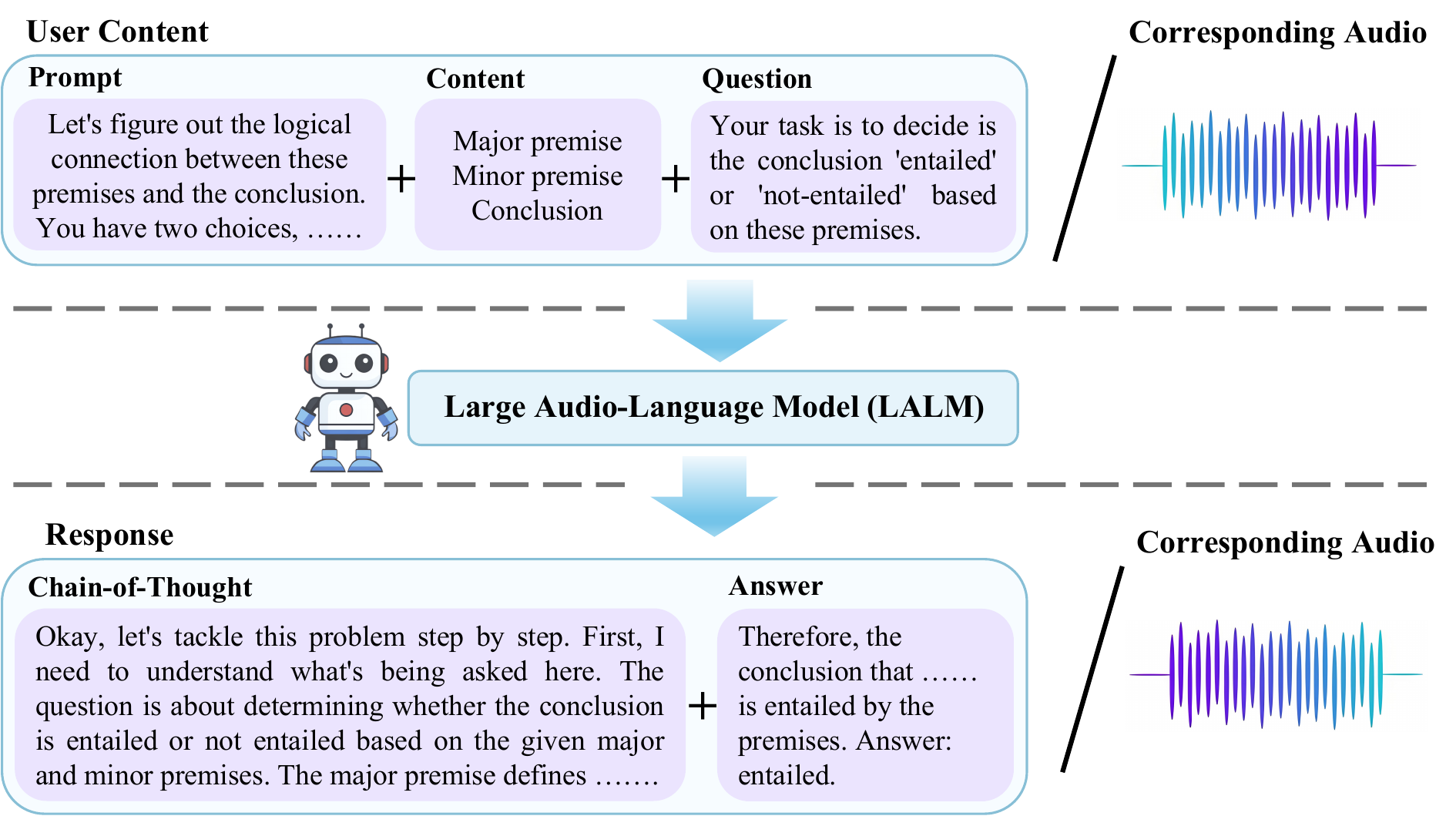}
\caption{Overview of a SoundMind sample for the audio logical reasoning (ALR) task. Each instance contains \textbf{User Content} (top), including a natural-language prompt, structured logical
triplets, and a question, as well as \textbf{Response} (bottom), which consists of step-by-step chain-of-thought reasoning and the final answer. Both components are provided in text and accompanied by the corresponding input–output speech audio (right). SoundMind therefore offers \textbf{complete logical reasoning tasks} with fully aligned content and response across text and audio modalities, supporting multimodal training and evaluation of large audio–language models.}
  \label{fig:task}
\end{figure*}

Large language models (LLMs) have made remarkable strides in reasoning capabilities through innovations such as Chain-of-Thought (CoT) prompting and specialized reasoning architectures. Recent models such as OpenAI's o1 model \citep{jaech2024openai} and Deepseek-R1 \citep{guo2025deepseek} have demonstrated exceptional performance on complex logical tasks \citep{team2025kimi, zhao2024marco,yuan2025superficial,zhang2024working,zhang2025overcoming,zhang2025growing,han2024infimm}. Particularly noteworthy is Deepseek-R1's rule-based reinforcement learning approach, which enables emergent reasoning without relying on traditional frameworks like Monte Carlo Tree Search \citep{wan2024alphazero} or process reward models \citep{lightman2023let}. This general reasoning paradigm has been successfully extended to the visual domain, where frameworks such as Visual-CoT \citep{shao2024visual} have significantly improved multimodal models' cognitive abilities for image and video reasoning.

Despite these advances in text and visual reasoning, there is a significant gap in \textit{audio reasoning and generation capabilities}. While large audio-language models (LALMs) like Audio Flamingo \citep{kong2024audio} and Qwen2.5-Omni-7B \citep{xu2025qwen2} have made progress in audio understanding, end-to-end audio reasoning remains underdeveloped. This limitation stems primarily from two factors: (1) the simplicity of existing audio reasoning datasets \citep{suzgun2023challenging,kong2024audio}, which often contain only brief textual labels without proper audio modality annotations, and (2) the technical challenges in maintaining reasoning coherence during long-duration audio generation. Current CoT methods applied to audio often lead to hallucinations and performance degradation when generating extended reasoning sequences. The lack of aligned audio-text annotations further impedes research on reasoning-driven audio generation, creating a bottleneck in developing LALMs with sophisticated reasoning capabilities. 

To address these challenges in audio reasoning, we introduce a comprehensive approach with two key components. First, we introduce SoundMind, a benchmark dataset specifically designed for the audio logical reasoning (ALR) task. It contains 6,446 audio–text aligned samples, each containing user content, step-by-step chain-of-thought reasoning, final answers, and the corresponding input–output speech audio (Figure \ref{fig:task}). Built upon the LogiQA 2.0–NLI dataset~\citep{liu2023logiqa}, SoundMind preserves the full logical structure and augments it with parallel text–audio annotations, enabling models to learn reasoning grounded in audio. Second, we propose SoundMind-RL, a rule-based reinforcement learning algorithm that addresses the challenges of long-form audio reasoning generation. Drawing inspiration from the Logic-RL framework \citep{xie2025logic}, SoundMind-RL incorporates a strict format-based reward mechanism to prevent shortcut reasoning biased toward the textual modality. It leverages the REINFORCE++ algorithm \citep{hu2025reinforce++} alongside the reward design principles from Deepseek-R1 for effective post-training. Our main contributions are summarized as follows:
\begin{itemize}[leftmargin=*]
    \item \textbf{SoundMind Dataset:} We release a high-quality audio logical reasoning dataset comprising 6,446 samples with user content, CoT reasoning, answers and corresponding audio input/output. It provides deep reasoning annotations in both text and audio modalities, serving as a valuable resource for RL-based training and evaluation of audio–language models.
    \item \textbf{SoundMind-RL Algorithm:} Taking into account the unique challenges of audio reasoning generation, we design a strict format-based reward mechanism that maintains reasoning coherence across modalities. By combining the improved REINFORCE++ algorithm with reward design principles from DeepSeek-R1, we fine-tune Qwen2.5-Omni-7B through reinforcement learning, achieving state-of-the-art results across three input–output modality configurations on the SoundMind benchmark.
\end{itemize}

\begin{table*}[t]
  \centering
  \resizebox{1.0\textwidth}{!}{\begin{tabularx}{\textwidth}{lccccc}
    \toprule
    \textbf{Datasets} & \textbf{Type} & \textbf{Transcripts} & \textbf{CoT annotations} & \textbf{Samples} & \textbf{Hours} \\
    \midrule
    CoTA              & Sound, Speech, Music & \ding{56} & \ding{52} & 1.2M & 6K        \\
    AudioSkills    & Speech & \ding{56} & \ding{56} & 4.2M & 9.3K    \\
    LongAudio    & Audio & \ding{56} & \ding{56} & 263K & 8.5K    \\
    Big Bench Audio & Speech & \ding{56} & \ding{56} & 1000 & 2 \\
    \midrule
    \textbf{SoundMind (Ours)} & Speech & \ding{52} & \ding{52} & 6446 & 1K \\
    \bottomrule
  \end{tabularx}}
\caption{Comparison of publicly available datasets for audio reasoning. Existing resources differ in scope: some provide speech data without transcripts, while others include transcripts but lack step-by-step reasoning annotations aligned with audio. These gaps make systematic study of multimodal reasoning challenging. \textbf{SoundMind} combines transcripts, chain-of-thought (CoT) annotations, and speech audio into a unified benchmark, offering a balanced resource for training and evaluating audio–language models.}
\label{tab:compara}
\end{table*}

\section{Related Work}
\paragraph{Open-Source Datasets for Audio Reasoning.}
Research on audio reasoning remains limited by the scarcity of publicly available benchmarks, making systematic evaluation and comparison difficult. Existing corpora typically treat audio as descriptive input paired with text, yet their supervision remains purely textual, providing final annotations exclusively in text form~\citep{suzgun2023challenging,kong2024audio,ghosh2025audio}.   More recently, a number of studies have begun to recognize the critical role of chain-of-thought (CoT) prompting in audio reasoning datasets~\citep{xie2025audio}, highlighting the importance of structured reasoning traces. To the best of our knowledge, SoundMind is among the first open-source resources where audio itself serves as the annotated modality for reasoning, providing parallel text–audio representations and enabling models to learn from and be evaluated on spoken reasoning traces.

\paragraph{Chain-of-Thought Reasoning.}
LLMs enhance their reasoning ability through in-context learning (ICL), which processes prompts within the surrounding context. CoT techniques further reinforce this capability. Prominent CoT approaches include Tree-of-Thought (ToT)~\citep{yao2023tree}, manually crafted few-shot CoT~\citep{wei2022chain}, and various automatic generation strategies~\citep{jin2024zero}. Recent works have also examined the necessity, theoretical underpinnings, and task-specific effectiveness of CoT reasoning~\citep{sprague2024cot}. OpenAI's ol model~\citep{jaech2024openai} has reignited interest in CoT prompting and is often paired with reinforcement learning-based training approaches~\citep{hu2025reinforce++,xie2025logic}.

\paragraph{Multimodal Chain-of-Thought Reasoning.}
CoT prompting has seen notable advancements in the multimodal domain. For instance, Visual-CoT~\citep{shao2024visual} integrates object detection to assist reasoning, while LLaVA-CoT~\citep{xu2024llava} and MAmmoTH-VL~\citep{guo2024mammoth} improve performance via dataset augmentation. However, CoT applications in the audio domain are still in their infancy. Audio-CoT~\citep{ma2025audio} demonstrates that zero-shot CoT prompting yields improvements on simple audio tasks, but remains inadequate for complex reasoning.

Although current audio-language models (ALMs) have made progress in comprehension and real-time response, their capability in CoT-style reasoning remains underexplored. Our study addresses this research gap by systematically investigating the application of CoT techniques within ALMs.

\section{Dataset Design and Construction}
\subsection{Key Characteristics}
We introduce SoundMind, a unified benchmark for the audio logical reasoning (ALR) task, specifically developed to advance research on reasoning grounded in auditory input. In contrast to most existing datasets that provide only textual supervision, SoundMind offers both text-level and audio-level chain-of-thought (CoT) annotations, allowing models to learn and reason directly from spoken prompts and responses. The dataset emphasizes speech-based scenarios, making it well suited for applications such as dialogue understanding and auditory commonsense reasoning.

As shown in Table~\ref{tab:compara}, existing audio datasets such as CoTA~\citep{xie2025audio}, AudioSkills~\citep{ghosh2025audio}, LongAudio~\citep{ghosh2025audio}, and Big Bench Audio~\citep{suzgun2023challenging} provide valuable resources but remain limited in their support for auditory reasoning. Some lack audio-level annotations entirely, while others include audio but omit chain-of-thought (CoT) reasoning traces or detailed step-by-step rationales. For instance, although CoTA offers CoT-style annotations, it does not include audio transcripts, which limits its utility for training and evaluating models that operate directly on speech input. Likewise, AudioSkills and LongAudio supply a large quantity of audio data but no reasoning supervision, making them less suitable for investigating step-by-step inference or developing models that require explicit alignment between input and reasoning output.

\begin{figure*}[t]
  \includegraphics[width=\textwidth]{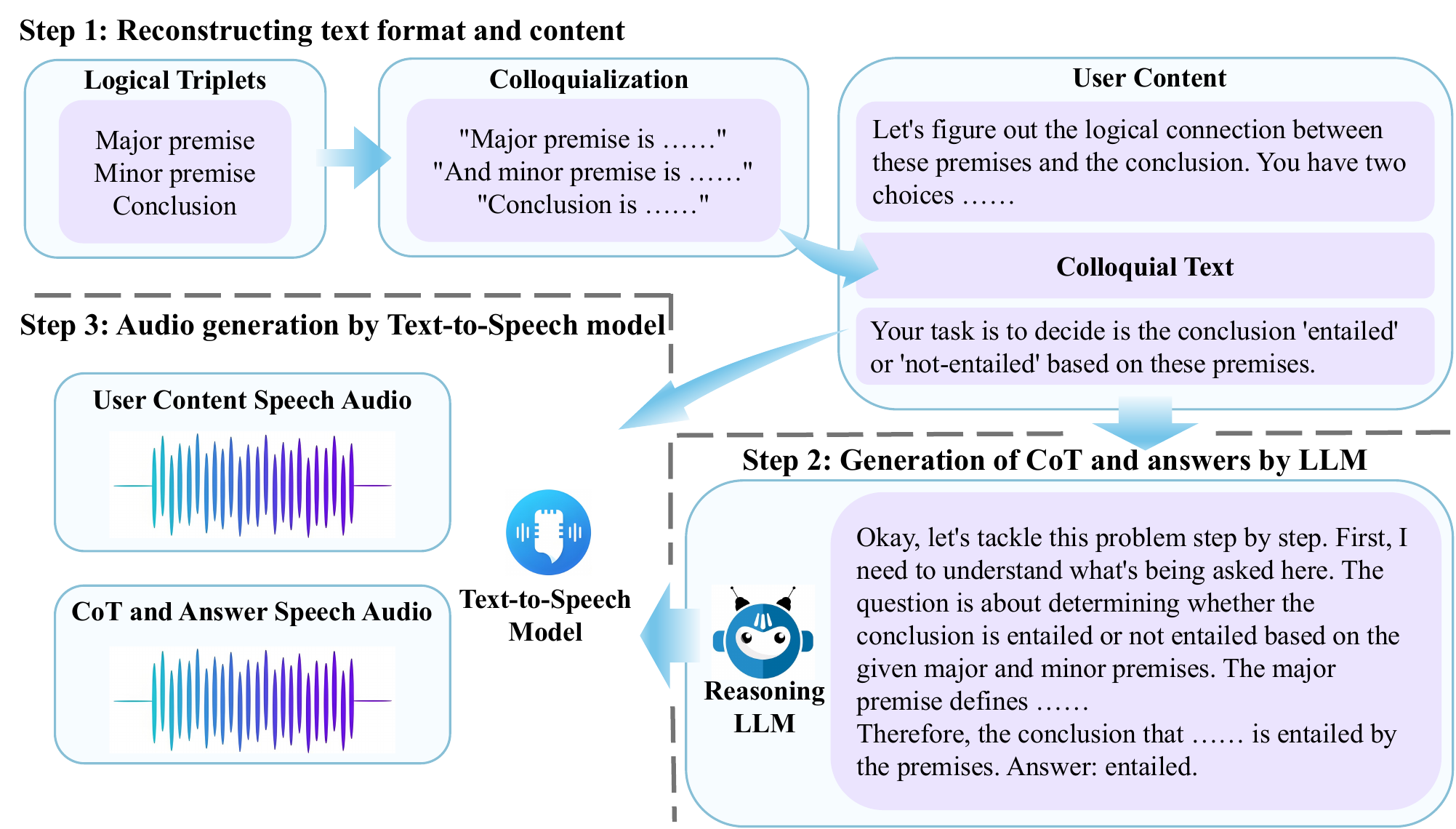}
\caption{Three-step pipeline for constructing SoundMind samples. 
\textbf{Step 1:} Structured logical triplets are converted into natural, conversational prompts through a Colloquialization module. 
\textbf{Step 2:} A large language model generates detailed chain-of-thought reasoning and final answers, providing rich supervision for reasoning tasks. 
\textbf{Step 3:} Both the user content and the generated reasoning–answer pairs are converted into speech using a text-to-speech (TTS) model, yielding two carefully aligned audio segments that together capture the full reasoning interaction.}
  \label{fig:step}
\end{figure*}

The SoundMind dataset is constructed through a structured pipeline that transforms textual logical reasoning tasks into natural spoken audio interactions, thereby enabling the development of audio-based reasoning models. The complete pipeline is illustrated in Figure~\ref{fig:step}.

Although SoundMind contains fewer samples (6,446) than large-scale datasets such as AudioSkills, it is specifically designed for logic-oriented reasoning and provides over 1,074 hours of carefully annotated speech audio with both text-level and audio-level CoT supervision. This comprehensive annotation makes SoundMind a valuable benchmark for training and evaluating models that require structured, multimodal reasoning grounded in audio perception.

\subsection{Data Generation Pipeline}

Our pipeline takes as input structured logical triplets consisting of a major premise, a minor premise, and a conclusion. These are first processed by a \textbf{Colloquialization} module that rewrites the formal statements into natural, conversational prompts (e.g., \textit{“The major premise is …”}, \textit{“Let’s figure out the logical connection …”}). This step enhances the fluency of the resulting audio and more closely reflects realistic user queries.

\renewcommand{\arraystretch}{1.2}
\begin{table*}[ht]
  \centering
  \small
  \begin{tabularx}{\textwidth}{>{\bfseries}p{3.2cm}|>{\raggedright\arraybackslash}X}
    \toprule
    Position & \textbf{Prompt} \\
    \midrule
    System & Your task is to decide if the conclusion is "entailed" or "not-entailed" based on these premises. You are a wise person who answers two-choice questions, "entailed" or "not entailed". Use plain text for thought processes and answers, not markdown or LaTeX. The thought process and response style should be colloquial, which I can then translate directly into audio using the TTS model. The final output is the Answer, nothing else, and the format is Answer: YOUR ANSWER. For example: "Answer: entailed." or "Answer: not entailed." The final answer must contain nothing else! The thought process should be very complete, careful, and cautious. When you think and generate a chain of thought, you need to test your answer from various angles. \\
    \midrule
    Before the major premise &
    Let's figure out the logical connection between these premises and the conclusion. You have two choices: "entailed" means the conclusion must be true based on the given premises, or "not-entailed" means the conclusion can't be true based on the premises. Here's the setup: \\
    \midrule
    Behind the conclusion &
    Your task is to decide is the conclusion is "entailed" or "not-entailed" based on these premises. \\
    \bottomrule
  \end{tabularx}
  \caption{Instructional prompts used at different positions during chain-of-thought generation. 
The \textit{system} prompt specifies the overall task, output format, and encourages a careful, colloquial reasoning style suitable for speech synthesis. 
Additional \textit{contextual prompts} placed before the major premise and after the conclusion provide further guidance, helping the model interpret logical relations and produce well-structured reasoning traces.}
  \label{tab:prompt}
\end{table*}

\begin{figure*}[ht]
  \includegraphics[width=\textwidth]{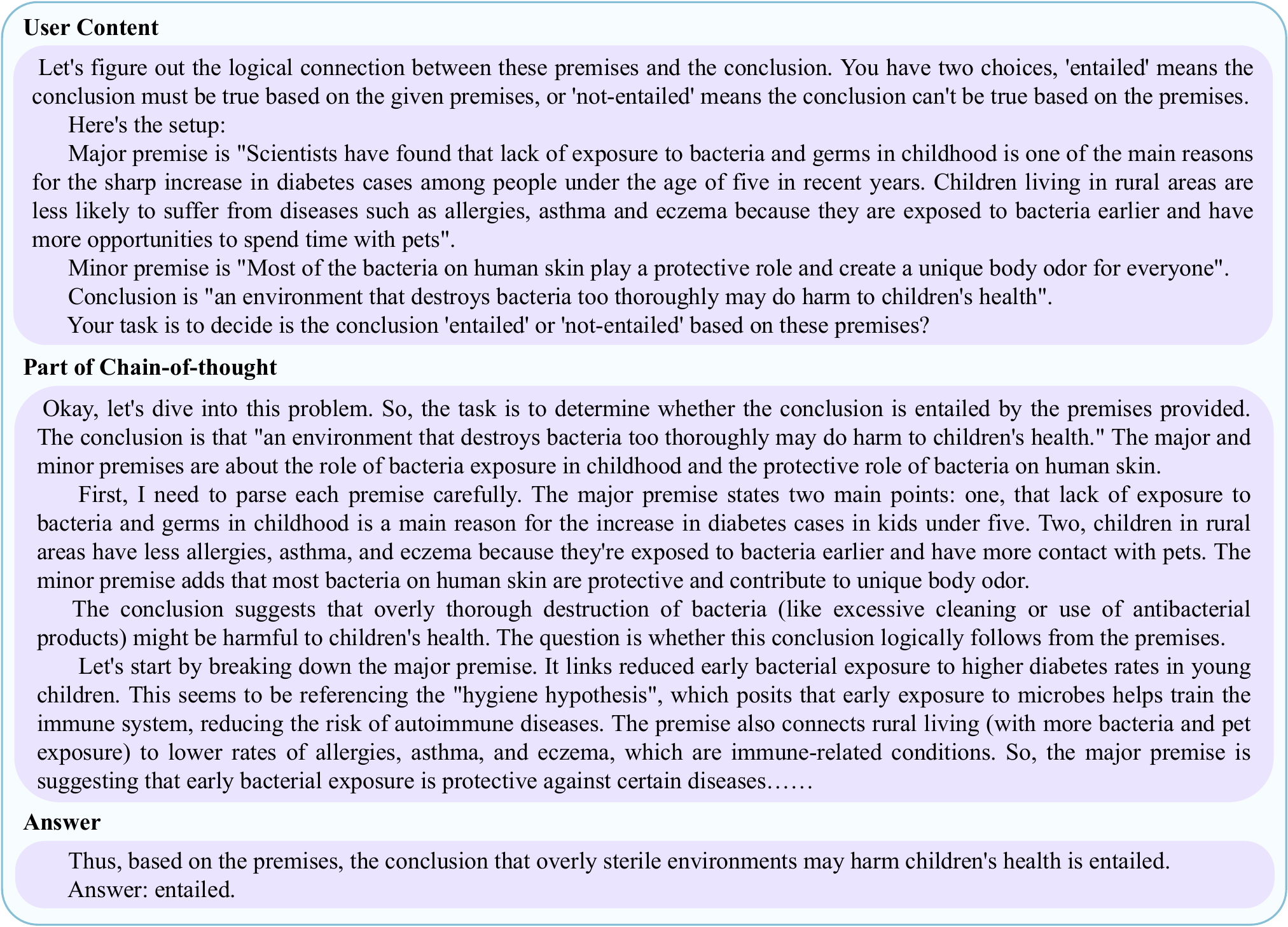}
  \caption{An illustrative example from the SoundMind dataset. Each sample contains three components: 
(1) \textbf{User Content}, which includes a natural-language prompt and a structured logical triplet (major premise, minor premise, conclusion); 
(2) \textbf{Chain-of-Thought}, a step-by-step reasoning trace generated by a large language model, explaining the entailment decision; 
and (3) the final \textbf{Answer}. 
All components are provided in both text form and synthesized speech audio, enabling fully multimodal training and evaluation of audio–language models.}
  \label{fig:sample}
\end{figure*}

The colloquialized content and instructions are merged into a single \textbf{User Content} block and synthesized into speech audio using the high-fidelity text-to-speech model MegaTTS 3~\citep{jiang2025megatts}, forming the input side of the dataset. The prompts used in this process are listed in Table~\ref{tab:prompt}.

To generate the output, we employ the large language model DeepSeek-R1 to produce step-by-step CoT reasoning and final answers, providing richer supervision for model training. The generated reasoning and answers are then synthesized into speech audio with MegaTTS 3~\citep{jiang2025megatts}, yielding the corresponding spoken response.

As a result, each SoundMind instance consists of two carefully aligned audio segments: one representing the user query and the other presenting the detailed step-by-step reasoning and the corresponding final answer, as illustrated in Figure~\ref{fig:sample}.

\section{SoundMind-RL Algorithm}
\label{sec:method}
We improve our system through iterative optimization of rule-based rewards. All $\lambda$ in the following equations are hyperparameters.

\paragraph{Answer Format Correctness Evaluation.}
To ensure the correctness of answer formatting, we first require that the token ``Answer:'' must appear within the last five characters of the model response. Considering that the SoundMind dataset covers both text and audio modalities, we design two specific format scoring methods (denoted as $S_{\text{format}}$), calculated as follows:
\begin{equation}
\mathit{S}_{\text{format}}^{(1)} =
\begin{cases}
\lambda_1, & \text{if correct for text tokens} \\
0, & \text{otherwise},
\end{cases}
\end{equation}
\begin{equation}
\mathit{S}_{\text{format}}^{(2)} =
\begin{cases}
\lambda_2, & \text{if correct for audio tokens} \\
0, & \text{otherwise}.
\end{cases}
\end{equation}

\paragraph{Answer Correctness Evaluation.}
Once the format compliance is verified, this module evaluates the factual accuracy of the model response. Specifically, the answer score ($S_{\text{answer}}$) is computed based on the consistency between the model's predicted answer and the ground truth answer, using the following formulation:
\begin{equation}
\mathit{S}_{\text{answer}} = 
\begin{cases}
\lambda_3, & \text{if answer = ground truth} \\
0, & \text{otherwise}.
\end{cases}
\end{equation}

\paragraph{Reasoning Length Evaluation.}
We additionally introduce a reward evaluation based on the reasoning length of the model response. This score ($S_{\text{len}}$) is computed by comparing the ratio of the model output length to the reference reasoning length. Given that the SoundMind dataset provides supervision in both text and audio modalities, we design two distinct length evaluation methods, defined as follows:
\begin{equation}
\mathit{S}_{\text{len}}^{(1)} = 
\lambda_{4}\times\min\left(1, \frac{L_{\text{model}}}{L_{\text{annotation}}} \right),
\end{equation}
\begin{equation}
\mathit{S}_{\text{len}}^{(2)} = 
\lambda_{5}\times\min\left(1, \frac{T_{\text{model}}}{T_{\text{annotation}}} \right).
\end{equation}
where $L$ denotes the length of text tokens, and $T$ denotes the length of audio tokens.

\paragraph{REINFORCE++ Policy Optimization.}
In our setting, the policy $\pi_\theta$ corresponds to a large-scale ALM, which receives an audio question as input and produces a reasoning response in either text or audio form. To optimize the model with the above composite reward, we adopt the REINFORCE++~\cite{hu2025reinforce++}—a clipped policy-gradient method that eliminates the need for a value (critic) network while leveraging PPO-style stability and sample efficiency.
Specifically, REINFORCE++ updates the policy by maximizing the following objective:
\begin{equation}
\begin{aligned}
& \mathcal{J}_{\mathrm{REINFORCE++}}(\theta) =\;\mathbb{E}_{(x, y) \sim \mathcal{D},\; o_{\leq t} \sim \pi_{\theta_\mathrm{old}}}
\\
& \Bigg[\min\left(
        r_t(\theta)\, \hat{A}_t,\;
        \operatorname{clip}\left(r_t(\theta), 1-\epsilon, 1+\epsilon\right)\, \hat{A}_t
    \right)
\Bigg]
\end{aligned}
\end{equation}
where $x$ is the audio input, $y$ is the generated response, and $r_t(\theta) = \frac{\pi_\theta(a_t|s_t)}{\pi_{\theta_\mathrm{old}}(a_t|s_t)}$ is the importance sampling ratio. $\hat{A}_t$ is the normalized advantage, given by:
$
A_t = R(x, y) - \beta \sum_{i=t}^{T} \mathrm{KL}(i),  
\hat{A}_t = \frac{A_t - \mu_A}{\sigma_A}.
$
Here, $R(x, y)$ is the cumulative reward, composed of the weighted sum of the above reward terms ($S_{\text{format}}$, $S_{\text{answer}}$, $S_{\text{len}}$), and $\mathrm{KL}(i)$ denotes the token-level KL divergence between the current policy and the reference SFT model. The KL term is defined as:
\begin{equation}
\mathrm{KL}(i) = \log \frac{\pi_{\theta}(a_i|s_i)}{\pi_{\mathrm{SFT}}(a_i|s_i)},
\end{equation}
where $\pi_{\theta}$ is the current policy being optimized, $\pi_{\mathrm{SFT}}$ is the fixed reference (supervised fine-tuned) model, $\beta$ is the KL penalty coefficient, and $\mu_A$, $\sigma_A$ are the batch mean and standard deviation for normalization. $\epsilon$ is the PPO clip parameter.
REINFORCE++ combines the stability of PPO's clipped surrogate loss with the efficiency of critic-free Monte Carlo policy gradient updates, using a token-level KL penalty and batch-normalized advantages. It trains audio-language models for end-to-end reasoning over audio questions.

\section{Experiments}

\begin{table*}[t]
  \centering
    \resizebox{1\textwidth}{!}{\begin{tabular}{lcccccc}
  % \begin{tabularx}{\textwidth}{Y|YYYYYY}
    \toprule
\textbf{Split} & \textbf{\# Entailed} & \textbf{\# Not-ent.} & \textbf{Avg. Inp. Tok.} & \textbf{Avg. Out. Tok.} & \textbf{Avg. Inp. Dur. (s)} & \textbf{Avg. Out. Dur. (s)} \\ \midrule
    Training   & 2326 & 2858 & 182 & 1683 & 62.33  & 608.42      \\
    Test    & 296 & 360 & 158 & 1424 & 57.90 & 586.51    \\
    Validation    & 264 & 342 & 155 & 1426 & 57.85 & 586.39    \\
    \bottomrule
  \end{tabular}}
\caption{ Detailed statistics of the SoundMind dataset across training, validation, and test splits. “\# Entailed” and “\# Not-ent.” denote the number of samples labeled as entailed and not-entailed. “Avg. Inp./Out. Tok.” reports the mean token counts for the input and outputs, respectively. “Avg. Inp./Out. Dur. (s)” provides the average duration (in seconds) of the input audio and the corresponding output audio. Together, these statistics highlight the dataset’s scale, near-balanced class distribution, and focus on long-form reasoning supervision.}
  \label{tab:alrsta}
\end{table*}

\begin{figure*}[t]
    \centering
    \begin{subfigure}[t]{0.32\textwidth}
        \includegraphics[width=\linewidth]{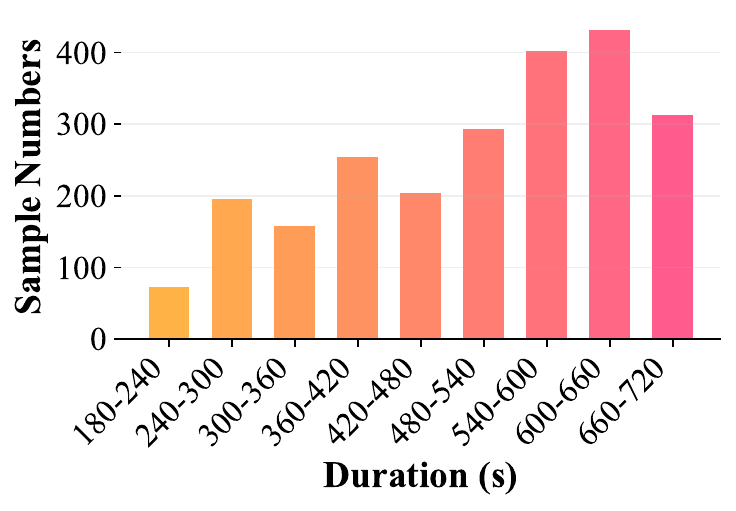}
        \caption{Training set (Entailed)}
        \label{fig:sa}
    \end{subfigure}
    \hfill
    \begin{subfigure}[t]{0.32\textwidth}
        \includegraphics[width=\linewidth]{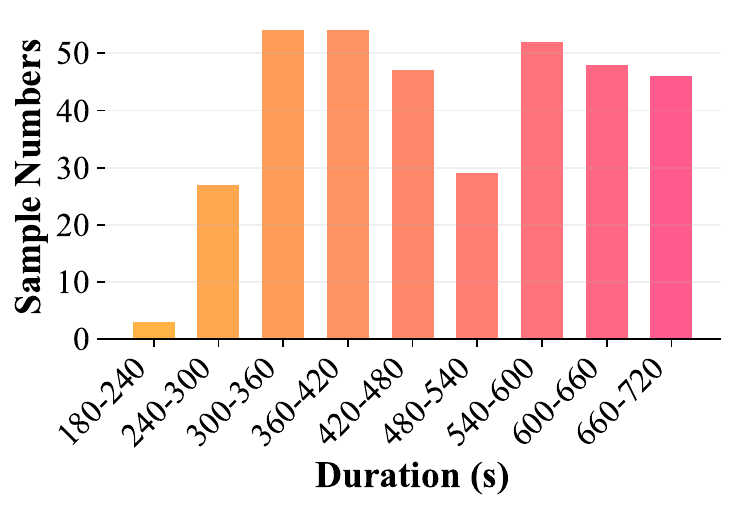}
        \caption{Test set (Entailed)}
        \label{fig:sb}
    \end{subfigure}
    \hfill
    \begin{subfigure}[t]{0.32\textwidth}
        \includegraphics[width=\linewidth]{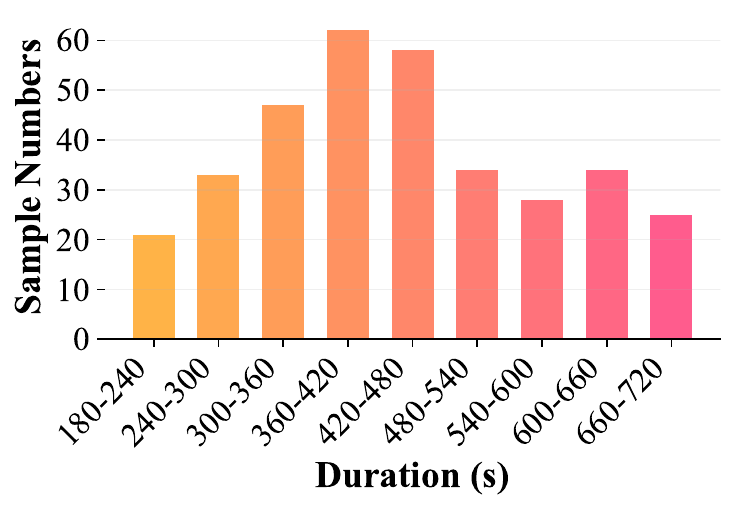}
        \caption{Validation set (Entailed)}
        \label{fig:sc}
    \end{subfigure}
    
    \begin{subfigure}[t]{0.32\textwidth}
        \includegraphics[width=\linewidth]{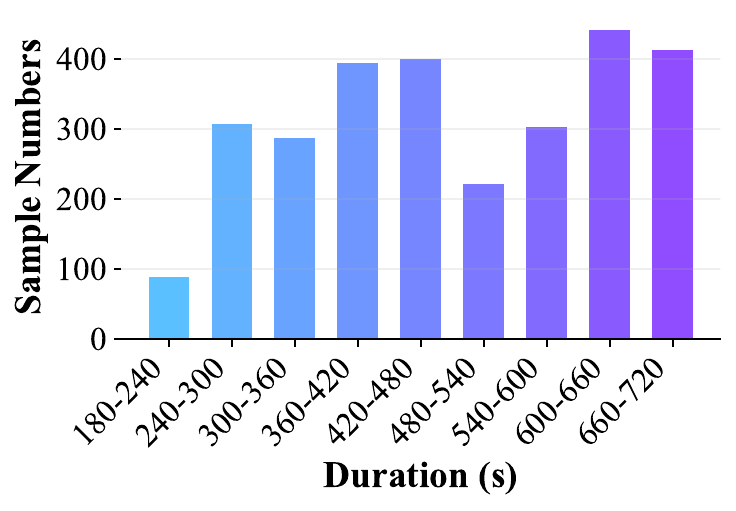}
        \caption{Training set (Not-entailed)}
        \label{fig:sd}
    \end{subfigure}
    \hfill
    \begin{subfigure}[t]{0.32\textwidth}
        \includegraphics[width=\linewidth]{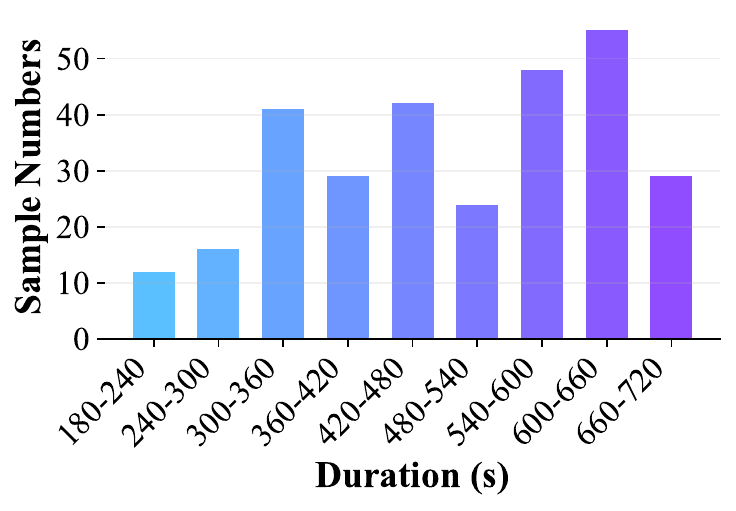}
        \caption{Test set (Not-entailed)}
        \label{fig:se}
    \end{subfigure}
    \hfill
    \begin{subfigure}[t]{0.32\textwidth}
        \includegraphics[width=\linewidth]{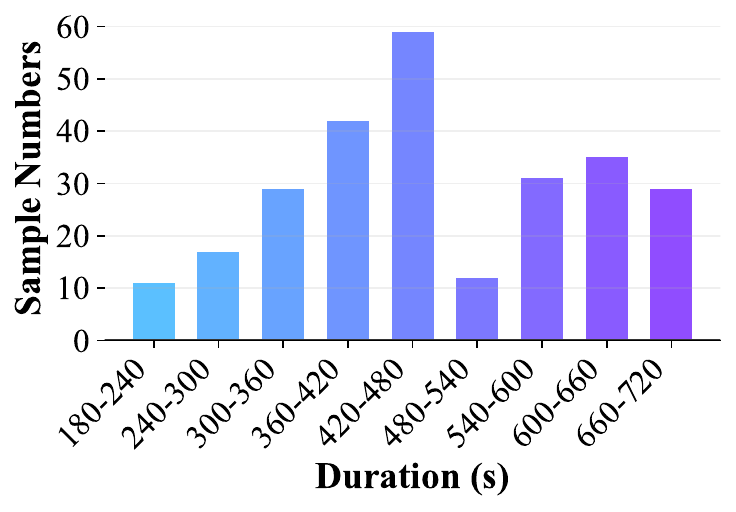}
        \caption{Validation set (Not-entailed)}
        \label{fig:sf}
    \end{subfigure}
\caption{
Distribution of audio durations across dataset splits and entailment labels. 
Subfigures (a–c) show histograms for “Entailed” samples in the training, test, and validation sets, 
while (d–f) show the corresponding distributions for “Not-entailed” samples. 
These plots illustrate the broad coverage of audio durations and the consistent duration profiles across splits, 
highlighting that the SoundMind dataset offers both diversity and balance for evaluating model performance across different temporal contexts.
}
    \label{fig:stat}
\end{figure*}

\subsection{Setup}
We fine-tune the Qwen2.5-Omni-7B model on the SoundMind dataset using the proposed SoundMind-RL algorithm. All experiments were conducted under a consistent hardware environment, consisting of an Intel(R) Xeon(R) Platinum 8468 CPU and 8 NVIDIA H800 GPUs, each equipped with 80 GB of memory.
The weighting coefficients for the reward components were set as follows: $\lambda_1 = 1.0$, $\lambda_2 = 0.5$, $\lambda_3 = 2.0$, $\lambda_4 = 1.0$, and $\lambda_5 = 0.75$. The training procedure was executed for 50{,}000 steps.
All other hyperparameter settings followed those used in Logic-RL~\cite{xie2025logic}.

To fully exploit the availability of both text and audio modalities in the SoundMind dataset, 
we evaluate our approach under three input–output configurations:  
\circled{1} Table~\ref{tab:a2t}: audio-only input with text-based reasoning output, 
benchmarked against five strong multimodal LLMs: MiniCPM-o~\citep{yao2024minicpm}, Gemini-Pro-V1.5~\citep{team2024gemini}, 
Baichuan-Omni-1.5~\citep{li2025baichuan}, Qwen2-Audio~\citep{chu2024qwen2}, and Qwen2.5-Omni-7B~\citep{xu2025qwen2}.  
\circled{2} Table~\ref{tab:t2a}: text-only input with audio-based reasoning output. 
Due to the lack of comparable open-source multimodal models, we use Qwen2.5-Omni-7B as the reference baseline.  
\circled{3} Table~\ref{tab:a2a}: audio-only input with audio-based reasoning output. Together, these three configurations allow us to comprehensively assess SoundMind-RL across text and audio modalities, covering fully cross-modal reasoning scenarios.

\subsection{Dataset Analysis}
Table~\ref{tab:alrsta} summarizes key statistics of the SoundMind dataset, clearly illustrating its overall scale, class balance, and linguistic diversity across training, validation, and test splits. 
The corpus contains 6,446 samples with a near-balanced distribution of entailed and not-entailed instances, thereby ensuring representative coverage of both classes for robust entailment prediction.

The user content averages 160–180 tokens across all splits, whereas the generated CoT responses are much longer, exceeding 1,400 tokens on average. This substantial length gap reflects the dataset’s focus on detailed, step-by-step reasoning, providing rich supervision that encourages faithful and transparent reasoning in audio-based tasks.

In terms of audio representation, the user content speech averages roughly one minute per sample, whereas the CoT reasoning audio extends to nearly ten minutes on average. This substantial duration provides a challenging and realistic benchmark for assessing models’ ability to perform long-form audio comprehension and sustained logical reasoning.

As shown in Figure~\ref{fig:stat}, the SoundMind dataset covers a broad range of audio durations, capturing important variability for developing and evaluating models for audio-based logical reasoning. 
Most segments fall within 3–12 minutes, with 28.6\% of training samples concentrated in the 540–720s range, which provides ample contextual information while keeping computational cost tractable. 
Notably, the dataset deliberately excludes very short clips to ensure that each sample contains sufficiently meaningful reasoning content.

The dataset exhibits a class distribution of 44.9\% entailed and 55.1\% not-entailed instances overall, closely resembling patterns in many real-world reasoning scenarios. This skew becomes more pronounced in the mid-range durations (240–420s), suggesting that longer and acoustically richer segments may introduce additional difficulty for reliable entailment classification. Notably, the 480–540s range in the training split shows a reversed distribution, with 56.9\% entailed samples, potentially reflecting distinctive acoustic patterns or systematic properties of the original source data.

The SoundMind dataset is split to maintain consistent entailment ratios across training (42.8\%), test (54.9\%), and validation (56.3\%) sets, thereby enabling reliable evaluation while preserving the inherent task complexity. Each split carefully preserves proportional coverage across all duration bins, effectively minimizing potential duration-related bias and ensuring representative diversity. The test set is nearly balanced (360 entailed vs. 296 not-entailed), providing a solid basis for rigorous and fair model assessment.

Overall, this distribution profile aligns with the dataset’s goal of advancing audio–language reasoning research by offering diverse but systematically controlled interaction scenarios. 
Such a design compels models to acquire robust cross-modal reasoning skills, discouraging reliance on superficial duration-based correlations and encouraging deeper semantic understanding.

\subsection{Results and Analysis}

\subsubsection{Audio-to-Text Reasoning}
\begin{table}[ht]
  \centering
  \resizebox{0.49\textwidth}{!}{\begin{tabular}{lc}
    \toprule
    \textbf{Model} & \textbf{Accuracy (\%)$\uparrow$}\\
    \midrule
    MiniCPM-o   & 73.17\\
    Gemini-Pro-V1.5    & 74.54\\
    Baichuan-Omni-1.5    & 70.58\\
    Qwen2-Audio    & 58.23\\
    Qwen2.5-Omni-7B    & 77.59\\
    \midrule
    \textbf{Qwen2.5-Omni-7B (SoundMind-RL)}  & \textbf{81.40}\\
    \bottomrule
  \end{tabular}}
\caption{
Accuracy (\%) of evaluated models on the SoundMind benchmark under the audio-to-text configuration. In this setting, models receive audio-only inputs and produce textual outputs containing the final answer. The table compares several strong multimodal LLM baselines with the reinforcement-tuned Qwen2.5-Omni-7B (SoundMind-RL).
}
  \label{tab:a2t}
\end{table}

Qwen2.5-Omni-7B fine-tuned with SoundMind-RL achieves state-of-the-art performance on the challenging audio-to-text reasoning task.  
Table~\ref{tab:a2t} reports results for generating textual outputs from audio-only inputs. Our model attains an accuracy of 81.40\%, clearly outperforming all evaluated baselines. Compared to its counterpart Qwen2.5-Omni-7B (77.59\%), the reinforcement-tuned variant yields an absolute improvement of 3.81\%, clearly demonstrating that the SoundMind reward framework substantially enhances the model's ability to produce accurate text-based logical conclusions from speech audio input.

Among the remaining baseline models, Gemini-Pro-V1.5 reaches 74.54\%, followed closely by MiniCPM-o (73.17\%) and Baichuan-Omni-1.5 (70.58\%). 
Qwen2-Audio performs the weakest, achieving only 58.23\%, clearly underscoring that generic multimodal alignment is insufficient for reasoning-intensive tasks without specialized optimization. 
Overall, Qwen2.5-Omni-7B (SoundMind-RL) further improves upon Gemini-Pro by 6.86\% and surpasses MiniCPM-o by 8.23\%, establishing a consistently clear lead across models of comparable or larger capacity. 
These findings strongly confirm that our reinforcement learning approach strengthens reasoning accuracy and generalizes effectively across audio reasoning tasks.

\subsubsection{Text-to-Audio Reasoning}

\begin{table}[ht]
  \centering
  \resizebox{0.49\textwidth}{!}{\begin{tabular}{lcc}
    \toprule
    \textbf{Model} & \textbf{WER (\%)$\downarrow$} & \textbf{Acc. (\%)$\uparrow$}\\
    \midrule
    Qwen2.5-Omni-7B   & \textbf{2.18} & 80.79\\
    \textbf{Qwen2.5-Omni-7B (SoundMind-RL)}  & 6.99 & \textbf{83.84}\\
    \bottomrule
  \end{tabular}}
\caption{Performance of Qwen2.5-Omni-7B and its SoundMind-RL fine-tuned variant on the SoundMind benchmark under the text-to-audio  setting. In this configuration, models receive text-only inputs and generate audio outputs. Results are reported using Word Error Rate (WER) to assess speech fidelity and accuracy (\%) to measure reasoning correctness.}
\label{tab:t2a}
\end{table}

SoundMind-RL significantly improves reasoning performance in the text-to-audio setting.  
As shown in Table~\ref{tab:t2a}, Qwen2.5-Omni-7B fine-tuned with SoundMind-RL achieves an accuracy of 83.84\%, yielding a 3.05\% improvement over the Qwen2.5-Omni-7B baseline.  
These results clearly indicate that the reward mechanism effectively encourages the model to produce coherent and consistent spoken reasoning, even without audio input.  
The observed accuracy gain is accompanied by a higher WER (6.99\% vs. 2.18\%), which can be attributed to the generation of longer and more detailed reasoning sequences under reinforcement learning, thereby increasing the chance of recognition errors while preserving semantic correctness.

\subsubsection{Audio-to-Audio Reasoning}

\begin{table}[ht]
  \centering
  \resizebox{0.49\textwidth}{!}{\begin{tabular}{lcc}
    \toprule
    \textbf{Model} & \textbf{WER (\%)$\downarrow$} & \textbf{Acc. (\%)$\uparrow$}\\
    \midrule
    Qwen2.5-Omni-7B   & \textbf{2.23} & 77.59\\
    \textbf{Qwen2.5-Omni-7B (SoundMind-RL)}  & 8.95 & \textbf{81.40}\\
    \bottomrule
  \end{tabular}}
\caption{Performance of Qwen2.5-Omni-7B and its SoundMind-RL fine-tuned variant on the SoundMind benchmark under the audio-to-audio reasoning setting. In this configuration, models receive audio-only inputs and generate audio reasoning outputs.  }
\vspace{0.1cm}
\label{tab:a2a}
\end{table}

In the fully audio-based setting, our SoundMind-RL framework delivers notable performance gains under the most challenging condition.  
Table~\ref{tab:a2a} presents results where models must derive and verbalize reasoning solely from acoustic input. Qwen2.5-Omni-7B fine-tuned with SoundMind-RL achieves 81.40\% accuracy, improving substantially over the baseline (77.59\%) and showing that the model can extract task-relevant logical dependencies directly from audio signals. The result is accompanied by a higher WER (8.95\% vs. 2.23\%), which we attribute to the generation of longer, more elaborate reasoning sequences and the added difficulty of learning from raw acoustic input. Our analysis attributes this to the absence of explicit fluency constraints during reinforcement learning, which instead prioritizes structured reasoning and semantic correctness.

\noindent\textbf{Takeaway.} Across all evaluated settings, SoundMind-RL consistently enhances logical reasoning performance for audio–language models.  Qwen2.5-Omni-7B fine-tuned with SoundMind-RL achieves substantial gains across audio-to-text, text-to-audio, and audio-to-audio configurations, outperforming all baseline models.  
These results demonstrate that reward-guided reinforcement learning can reliably improve multimodal reasoning accuracy by encouraging faithful answer formatting, factual correctness, and sufficiently detailed reasoning traces. For speech-based outputs, we observe a moderate increase in WER relative to the baseline, which correlates with the production of longer, information-rich reasoning segments and suggests that the model prioritizes semantic completeness over surface fluency.  
Future work can explore fluency-aware or prosody-sensitive reward components to better balance reasoning fidelity with the naturalness of generated speech.

\section{Conclusion}
We introduce SoundMind-RL, a novel rule-based reinforcement learning framework that empowers large-scale audio-language models with advanced logical reasoning capabilities across both audio and textual modalities. To support such training, we construct SoundMind, a dataset for the task of audio logical reasoning, containing 6,446 carefully curated audio–text pairs, each annotated with modality-specific chain-of-thought reasoning. Experimental results demonstrate that our method significantly improves performance and establishes state-of-the-art results on the SoundMind benchmark, consistently outperforming strong baselines across three reasoning settings: text-to-audio, audio-to-text, and audio-to-audio. We hope this work provides a useful resource and framework for advancing research in reasoning-oriented audio–language modeling and reinforcement-based multimodal learning.

\section*{Limitations}

While this study shows promising results, several limitations remain. (1) The rule-based reward design enforces reasoning format consistency but may introduce rigidity, and its generalization to more diverse or open-ended tasks remains to be examined. (2) The dataset relies on synthetic speech and automatically generated chain-of-thought annotations; despite careful curation, subtle artifacts or biases may still be present and could influence model behavior. (3) The increase in Word Error Rate (WER) observed in audio generation suggests a possible trade-off between reasoning depth and fluency, motivating further work on prosody modeling and decoding strategies. We leave these challenges for future work, aiming to advance the development of audio reasoning systems that are more robust, generalizable, and aligned with human needs.

\section*{Ethical Considerations}
We have not identified any ethical concerns directly related to this study.

\section*{Acknowledgment}
This study is supported by the Department of Defense grant HT9425-23-1-0267.

\bibliography{custom}

\appendix

\section{Baselines}
\label{baseline}
\begin{itemize}[leftmargin=*]
    \item {
    \textbf{MiniCPM-o} \cite{yao2024minicpm}: 
    An open-source multimodal language model supporting vision, speech, and real-time streaming. It emphasizes efficiency and deployability, making it suitable for interactive multimodal understanding and generation across a variety of general-purpose and resource-constrained scenarios.
    }
    \item {
    \textbf{Gemini-Pro-V1.5} \cite{team2024gemini}: 
    A multimodal model developed by Google that processes text, image, audio, and video inputs in a unified way. It is designed to handle extended context reasoning and cross-modal alignment, enabling coherent understanding and response in complex, multi-step tasks across modalities.
    }
    \item {
    \textbf{Baichuan-Omni-1.5} \cite{li2025baichuan}: 
    A multimodal model that combines text, vision, and audio understanding with audio generation. It aims to enable coherent cross-modal interaction and supports a wide range of multimodal and multitask applications in interactive settings.
    }
    \item {
    \textbf{Qwen2-Audio} \cite{chu2024qwen2}: 
    A speech language model that accepts both audio and text as input for natural voice-based interaction. It can perform detailed audio analysis, transcription, and multilingual understanding, serving as a general-purpose audio–text reasoning system.
    }
    \item {
    \textbf{Qwen2.5-Omni-7B} \cite{xu2025qwen2}: 
    A multimodal model that handles text, image, audio, and video inputs and generates both text and speech outputs. It supports streaming interactions and aims to provide robust multimodal reasoning and natural communication in real-time applications.
    } 
\end{itemize}

\section{Text-to-Speech Model}
\label{tts}
Constructing a fully human-recorded version of the audio logical reasoning dataset—covering thousands of user prompts, chain-of-thought reasoning traces, and final answers in both input and output modalities—would require extensive scripting, careful speaker coordination, and thousands of hours of professional narration. Such a large-scale recording effort is prohibitively expensive and logistically infeasible given the diversity and duration of reasoning sequences we target. Consistent with established practice in speech and multimodal learning research~\cite{jia2022leveraging,zevallos2022text,regmi2019nepali,cheng2024saic,bartelds2023making,Zhong2022ExternalTB}, we synthesize the audio modality using a high-fidelity text-to-speech (TTS) system. This strategy enables scalable and reproducible generation of paired input–output audio segments, while preserving fine control over speaking rate, prosody, and acoustic quality—key factors for training models to perform multi-minute reasoning without performance degradation.

To ensure that the synthesized audio faithfully reflects the complexity of logical reasoning, we conducted a systematic evaluation of several state-of-the-art open-source TTS systems. We compared MegaTTS-3~\cite{jiang2025megatts}, WhisperSpeech~\cite{WhisperSpeech}, Spark-TTS~\cite{wang2025spark}, and Zonos~\cite{zonos} across multiple dimensions, including perceived naturalness, prosody stability over long contexts, noise robustness, and intelligibility of extended reasoning chains. Each candidate system was benchmarked by synthesizing representative samples from our dataset and evaluated via multi-rater human listening studies. MegaTTS-3 consistently achieved the highest mean opinion scores, particularly in naturalness and prosodic consistency, qualities that we consider essential for faithfully preserving the chain-of-thought signal.

\begin{table*}[t]
  \centering
  \small
  \setlength{\tabcolsep}{6pt}
  \renewcommand{\arraystretch}{1.15}
  \resizebox{\textwidth}{!}{%
  \begin{tabular}{@{} l c c c c c | c c @{}}
    \toprule
    \textbf{Configuration} &
    $\lambda_1$ (TextFmt) &
    $\lambda_2$ (AudioFmt) &
    $\lambda_3$ (Answer) &
    $\lambda_4$ (TextLen) &
    $\lambda_5$ (AudioLen) &
    Accuracy (\%) &
    $\Delta$ vs. full (pp) \\
    \midrule
    (1) w/o audio rewards & 1.0 & 0   & 2.0 & 1.0 & 0    & 70.82 & -10.58 \\
    (2) w/o text rewards  & 0   & 1.0 & 2.0 & 0   & 1.0  & 48.84 & -32.56 \\
    (3) w/o answer reward & 1.0 & 0.5 & 0   & 1.0 & 0.75 & 60.24 & -21.16 \\
    (4) \textbf{full (Ours)} & 1.0 & 0.5 & 2.0 & 1.0 & 0.75 & \textbf{81.40} & --- \\
    \bottomrule
  \end{tabular}}
        \caption{Ablation of the composite reward in SoundMind-RL, evaluated on the test set. 
  $\lambda_1/\lambda_2$ enforce text/audio \emph{format} compliance, 
  $\lambda_3$ supervises \emph{answer correctness}, and 
  $\lambda_4/\lambda_5$ encourage \emph{reasoning length} in text/audio space. 
  Accuracy is reported in \%, and $\Delta$ indicates \emph{percentage-point (pp)} difference relative to the full model.}
  \vspace{-0.3cm}
      \label{tab:reward-ablation}
\end{table*}

On the basis of these findings, we select MegaTTS-3 as the synthesis engine for both user prompts and model reasoning traces. Its ability to produce long, coherent, and acoustically stable speech makes it well suited for the extended reasoning sequences characteristic of the ALR task. The resulting corpus provides a controlled yet realistic training signal that aligns with our objective of enabling robust, audio-native logical reasoning in large audio-language models. We further acknowledge the open-source speech community for providing high-quality TTS systems, which have been instrumental in enabling reproducible, large-scale construction of the audio datasets.

\section{Potential Applications}
\label{sec:applications}

Reasoning-capable ALMs can enable applications that demand explainability, transparency, and natural interaction.  
We highlight key domains where audio–native reasoning offers clear value and outline open challenges for future research.

\paragraph{Education and Tutoring.}
Spoken step-by-step explanations can help learners understand complex problems by exposing intermediate reasoning rather than presenting only a final answer \cite{wang2020step}.
This is particularly relevant for subjects such as logic, mathematics, and music, where reasoning over multimodal signals is essential for mastery.
Recent studies on music-related question answering and multimodal reasoning further highlight the need for models that can integrate audio and symbolic information to support learning and feedback \cite{you2025music,diao2025learning}.

\paragraph{Accessibility and Screen-Free Interaction.}
For users with visual impairments, hands-busy scenarios, or clinical settings, verbal reasoning offers an inclusive interface to access complex information. Replayable and summarizable reasoning can support comprehension and patient-centered decisions, consistent with efforts on trustworthy medical response generation \cite{li2024distinct} and healthcare voice assistants \cite{zhan2024healthcare}. Ensuring natural prosody and coherence over multi-minute speech remains crucial for high-stakes applications.

\paragraph{Conversational Agents and Voice Assistants.}
In applications such as troubleshooting, procedural guidance, and eligibility checking, users often ask “why” in addition to “what.” Models capable of structured, spoken justifications could reduce repeated follow-ups and improve user trust. 

\paragraph{Meetings, Lectures, and Knowledge Capture.}
Beyond transcription, audio–native reasoning models could verbalize intermediate inferences, summarize discussions, and surface entailment relations in real time, turning speech streams into structured knowledge. Robustness to overlapping speech, noise, and speaker variability remains a key challenge, as exemplified by the AMI Meeting Corpus~\cite{carletta2005ami}.

\section{Ablation Analysis}
\label{ablation}

\paragraph{Quantitative impact.}
Table~\ref{tab:reward-ablation} reports the results of ablating individual reward components.  
Removing \emph{text-format/length} supervision (Row~2) causes the most severe degradation, with accuracy dropping to 48.84\% (-32.56\% relative to the full model).  
This sharp decline underscores that structural guidance on the text side is essential for producing well-formed and logically verifiable reasoning traces.  
Excluding the \emph{answer-correctness} reward (Row~3) reduces accuracy to 60.24\% (-21.16\%), indicating that explicit factual supervision is necessary to ensure generated reasoning converges to the correct entailment decision rather than drifting into irrelevant but superficially coherent explanations.  
Removing \emph{audio-format/length} rewards (Row~1) still lowers accuracy to 70.82\% (-10.58\%), suggesting that modality-consistent formatting and duration constraints regularize the shared policy and enhance robustness.  
Overall, these results demonstrate that each reward term contributes meaningfully to model performance, and that their combination is crucial for maximizing reasoning accuracy.

\paragraph{Key takeaways.}
Table~\ref{tab:reward-ablation} highlights three main takeaways:  
(1) Text-side format and length supervision is the dominant factor for producing reliable, verifiable reasoning traces.  
(2) Answer-correctness reward is essential to align reasoning with ground truth labels and convert fluent explanations into correct decisions.  
(3) Audio-side rewards serve as powerful regularizers, improving policy robustness and benefiting reasoning quality across modalities.

\end{document}